\documentclass[11pt]{article}
\usepackage{cite}
\usepackage{amsmath,amssymb,amsfonts}
\usepackage{algorithmic}
\usepackage{graphicx}
\usepackage{textcomp}
\usepackage{xcolor}
\usepackage{url}
\usepackage{booktabs}
\usepackage[hidelinks]{hyperref}
\usepackage[a4paper,margin=1in]{geometry}
\usepackage{microtype}
\def\BibTeX{{\rm B\kern-.05em{\sc i\kern-.025em b}\kern-.08em
    T\kern-.1667em\lower.7ex\hbox{E}\kern-.125emX}}
\begin{document}

\title{LoRFT: Benchmarking Long-Range Vehicle Trajectory Reconstruction from Fixed Highway Cameras}

\author{Yufan Zhu, Kefu Yi\thanks{Corresponding author: \texttt{corfyi@csust.edu.cn}}, Xueju Zhang, Yunyang Tian,\\
Long Chen, and Zixuan Xiao\\[0.75em]
\normalsize School of Transportation, Changsha University of Science and Technology\\
\normalsize Changsha, China}
\date{}

\maketitle

\begingroup
\renewcommand{\thefootnote}{}
\footnotemark
\footnotetext{\footnotesize\raggedright
This work has been submitted to the IEEE for possible publication.
Copyright may be transferred without notice, after which this version may no longer be accessible.}
\addtocounter{footnote}{-1}
\endgroup

\begin{abstract}
Long-range vehicle trajectories provide important spatio-temporal evidence for traffic safety analysis, autonomous driving evaluation, and data-driven traffic management, but obtaining such records continuously from fixed highway cameras remains difficult. As vehicles move into distant road regions, perspective compression and scale decay cause automatic tracklets to fragment or terminate prematurely, even when the same vehicle can still be verified from motion continuity across neighboring frames. We formulate this setting as a spatio-temporal data mining problem: recovering the distant continuation of a vehicle trajectory from its reliable near-field tracklet. To support this task, we introduce LoRFT, to our knowledge the first open benchmark dedicated to long-range vehicle trajectory reconstruction from fixed highway cameras. LoRFT pairs reliable near-field tracklets with context-verified far-range references, and contains 22 expressway surveillance scenes, 366,109 video frames, 6,601 manually verified trajectories, 2,694,889 bounding boxes, road-geometry annotations, scene-level splits, and evaluation scripts. We further propose Map-RSTNet, a map-aware residual Seq2Seq model that reconstructs distant trajectories by anchoring sequence modeling in a road-geometry-aligned state space and dynamically refreshing local road geometry during decoding. On LoRFT, Map-RSTNet reduces the strongest baseline errors by 11.0\%, 15.4\%, and 10.5\% on ADE, FDE, and 5-second RMSE, respectively. These results indicate that road-geometry-aware reconstruction is a practical way to extend usable trajectory records from existing fixed-camera infrastructure. LoRFT further supports reproducible evaluation of trajectory mining, small-object detection, far-range vehicle tracking, and downstream traffic analysis. Processed trajectory annotations, road-geometry files, evaluation scripts, and source code are publicly available at \url{https://github.com/YvfanZhu/LoRFT}; the original surveillance videos are available under controlled access.
\end{abstract}

\noindent\textbf{Keywords:} vehicle trajectory reconstruction; fixed-camera highway surveillance; spatio-temporal data mining; road-geometry-aware modeling; trajectory benchmark.

\section{Introduction}

\begin{figure}[t]
\centering
\includegraphics[width=\columnwidth]{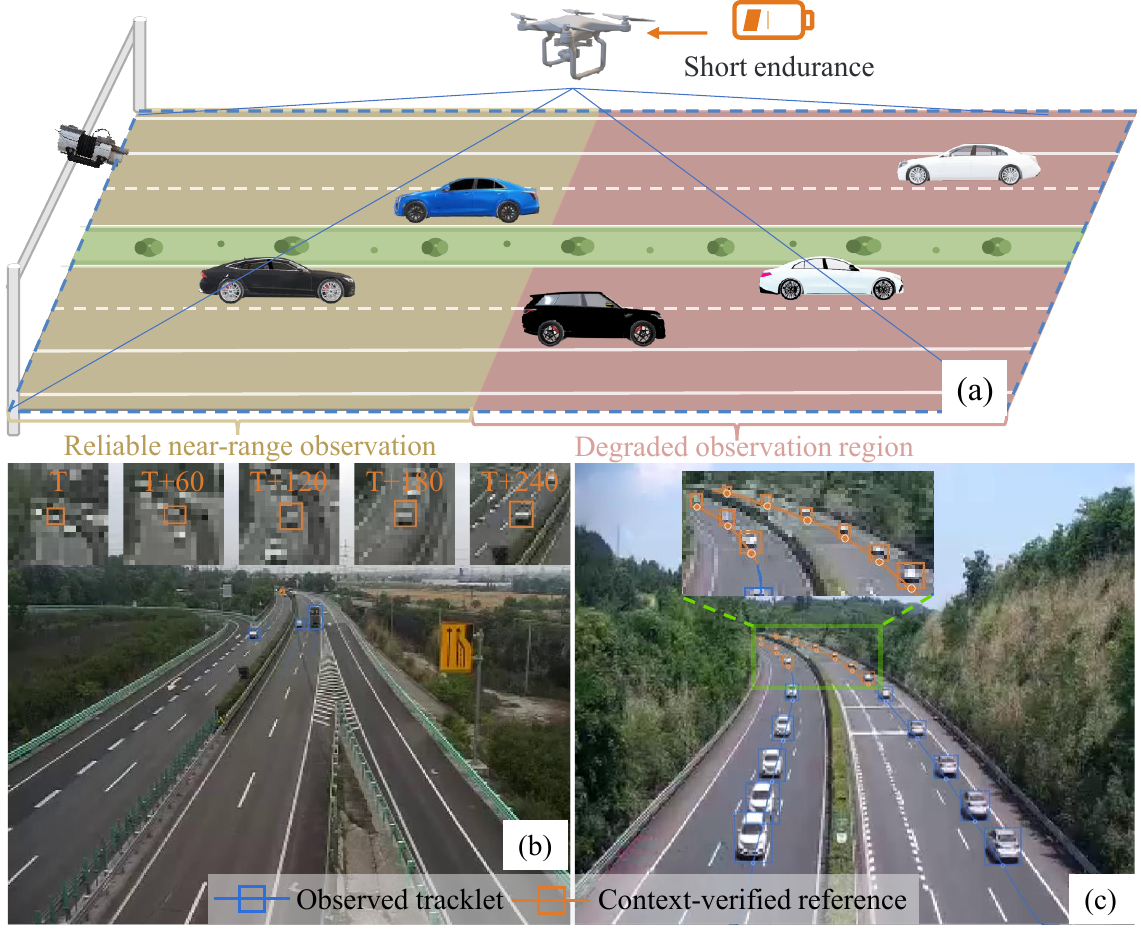}
\caption{Motivation and task definition of LoRFT.
(a) UAV sensing provides wide coverage but limited endurance; fixed cameras provide continuous monitoring but degraded far-range trajectory extraction.
(b) Far-range vehicles cannot be reliably identified in a single frame, but their identities can be verified across neighboring frames.
(c) LoRFT builds a reconstruction benchmark by pairing reliable near-field tracklets with context-verified far-range references from the same fixed-camera view.
Colored regions are illustrative rather than fixed spatial thresholds.}
\label{fig:motivation}
\end{figure}

Vehicle trajectory data provide essential spatio-temporal evidence for autonomous driving~\cite{itpnet,yang2025trajllm}, traffic safety analysis~\cite{hou2025conflict_prediction}, and data-driven traffic management~\cite{wu2024anisotropic,wang2024traffic_light}. Many of these applications depend on trajectory records with sufficient temporal span and spatial coverage to capture complete vehicle movements and the surrounding road context~\cite{wen2026hdsvt,i24_motion}. In highway scenarios, long-span trajectory collection commonly relies on aerial or overhead platforms and fixed highway cameras~\cite{highd,exid,fhwa_tmc_video,zou2022roadside}. As shown in Fig.~\ref{fig:motivation}(a), UAV-based sensing provides broad spatial coverage but is constrained by endurance, whereas fixed highway cameras support continuous monitoring but suffer from degraded far-range trajectory extraction. Fig.~\ref{fig:motivation}(b) further shows the observation pattern targeted in this paper: distant vehicles may be difficult to identify from a single frame, but their identities can still be verified from motion continuity across neighboring frames. This motivates the LoRFT task in Fig.~\ref{fig:motivation}(c), which reconstructs the far-range continuation of the same vehicle from a reliable near-field tracklet within one fixed-camera view.

The bottleneck in this setting is trajectory continuity rather than video acquisition. In fixed highway views, vehicles near the camera usually have sufficient image scale for detection and identity association, whereas distant vehicles are affected by perspective compression, scale decay, blurred appearance, and unstable association~\cite{kanhere2008lowangle,rope3d}. The same vehicle may remain visually identifiable from temporal context, but its automatically extracted tracklet can become fragmented, prematurely terminated, or missing. The reconstruction problem begins at this loss of continuity: the near-field tracklet anchors the vehicle identity and motion history, while the far-range continuation must be recovered after automatic tracking becomes unreliable. This framing differs from detecting distant vehicles independently, because the target remains the continuation of a known trajectory within the same fixed-camera view.

Highway trajectory datasets are abundant, but their annotations do not directly support the reconstruction problem described above. Overhead datasets such as highD and exiD provide accurate top-view motion records for traffic analysis and trajectory prediction~\cite{highd,exid}. Their sensing geometry, however, largely removes the distance-dependent scale decay, perspective compression, and identity fragmentation that appear in fixed surveillance videos. Fixed-camera datasets preserve the surveillance viewpoint, but they do not provide manual references for the far-range trajectory portions that automatically extracted tracklets fail to preserve. UA-DETRAC uses fixed traffic-camera videos but excludes low-resolution regions under its detection and tracking protocol~\cite{uadetrac}; CHD provides image-space highway trajectories, but these trajectories are mainly derived from detector-tracker outputs for prediction-oriented modeling~\cite{chd2024}. As a result, these representative datasets do not supply a manually verified continuation after the automatically extracted tracklet has become fragmented or prematurely terminated, even when the vehicle can still be traced from neighboring frames.

LoRFT manually annotates the far-range continuation that remains visible in the video after automatic tracking becomes unreliable. Starting from detector-tracker candidates, LoRFT retains the continuous and reliable segment of each selected vehicle as the observed input; when the automatic tracklet becomes fragmented or prematurely terminated in the distant road region, annotators follow the same vehicle across neighboring frames and verify its continuation as the reference segment. The resulting observed/reference pair is defined within one fixed-camera view: the model receives the trajectory portion that remains reliable after automatic extraction and verification, and is evaluated on the far-range continuation that remains visible in the video. LoRFT contains 22 expressway surveillance scenes from Sichuan Province, 366,109 video frames, and 6,601 manually verified vehicle trajectories, together with scene-level road-geometry annotations, predefined scene-level splits, and evaluation scripts. These annotations allow long-range reconstruction to be studied in the original fixed-camera image plane, under the perspective compression, scale decay, blurred far-range appearance, and association instability of routine highway monitoring.

These observed/reference pairs make the task evaluable, but the reconstruction itself must still account for how perspective changes vehicle motion in the image plane. In a fixed highway view, the apparent displacement and scale of a vehicle change with depth, while its feasible continuation follows the local road direction and boundaries. We therefore propose Map-RSTNet, which represents fixed-camera trajectories in a road-aligned state space before reconstructing the distant segment. The network uses geometry-aware residual decoding and projects the reconstructed trajectory back to the image plane for evaluation. Experiments on LoRFT show that this road-aligned reconstruction design reduces long-horizon errors compared with adapted trajectory modeling baselines.

The main contributions of this work are summarized as follows:

\begin{itemize}
\item We introduce LoRFT, to our knowledge the first open benchmark dedicated to long-range vehicle trajectory reconstruction from fixed highway cameras. LoRFT provides manually verified observed/reference trajectory pairs, vehicle bounding boxes, scene-level road-geometry annotations, predefined scene-level splits, evaluation scripts, and model code.

\item We propose Map-RSTNet, a map-aware residual Seq2Seq reconstruction network for recovering distant trajectory continuations in fixed surveillance views. The method represents trajectories in a road-aligned state space and uses geometry-aware decoding to reduce perspective-induced ambiguity during long-range reconstruction.
\end{itemize}

\section{Related Work}

\subsection{Vehicle Trajectory Datasets}

\begin{table}[t]
\centering
\caption{Comparison with representative highway-trajectory and traffic-camera datasets. 
T: trajectory; BB: bounding box; ``---'' denotes not reported or not applicable.}
\label{tab:dataset_comparison}
\scriptsize
\setlength{\tabcolsep}{2.6pt}
\renewcommand{\arraystretch}{1.12}
\begin{tabular*}{\columnwidth}{@{\extracolsep{\fill}}lcccccc}
\toprule
Dataset & View & Coord. & Scn. & FPS & Frames & Ann. \\
\midrule
NGSIM~\cite{ngsim}
  & Fixed-high & World/lane & 2 & 10 & 11.2K & T \\
UA-DETRAC~\cite{uadetrac}
  & Fixed-surv. & Image & 24 & 25 & 140K & 2D BB \\
highD~\cite{highd}
  & UAV-top & Metric & 6 & 25 & --- & T \\
exiD~\cite{exid}
  & UAV-top & Metric & 7 & 25 & --- & T \\
CHD~\cite{chd2024}
  & POV/high & Image & 8 & 5, 60 & 1.6M & T+2D BB \\
V2X-Seq~\cite{v2xseq}
  & Infra.-V2X & 3D/map & 95 & 10 & 15K & T+3D BB \\
\midrule
\textbf{LoRFT}
  & \textbf{Fixed-surv.} & \textbf{Image} & \textbf{22}
  & \textbf{25} & \textbf{366.1K} & \textbf{T+2D BB} \\
\bottomrule
\end{tabular*}
\end{table}

Vehicle trajectory datasets differ mainly in sensing viewpoint and observation completeness. NGSIM remains a widely used source for microscopic freeway traffic analysis \cite{ngsim}. Drone and overhead-view datasets provide more stable geometric observations: highD and exiD cover highway and ramp scenarios \cite{highd,exid}, while inD, rounD, openDD, and pNEUMA extend overhead trajectory collection to intersections, roundabouts, and urban traffic scenes \cite{ind,round,opendd,pneuma}. These datasets have supported trajectory modeling under comparatively complete observation conditions, but their viewpoints reduce image-space degradation factors that are common in fixed highway surveillance videos, such as target-scale decay, perspective distortion, and long-range track fragmentation.

Table~\ref{tab:dataset_comparison} summarizes the differences between LoRFT and representative highway-trajectory or traffic-camera datasets. NGSIM provides calibrated freeway trajectories in lane or world coordinates, while highD and exiD provide accurate overhead trajectories for highway and ramp scenarios~\cite{ngsim,highd,exid}. These datasets are useful for traffic analysis and trajectory prediction, but they do not retain the fixed-camera image-plane degradation targeted in LoRFT.

Traffic-camera datasets are closer in sensing setup but differ in annotation purpose. UA-DETRAC provides vehicle bounding boxes and identities for detection and multi-object tracking, and its protocol excludes image regions where vehicles cannot be reliably annotated~\cite{uadetrac}. CHD provides image-space highway trajectories, but its trajectories are mainly generated from detector-tracker outputs for prediction-oriented modeling~\cite{chd2024}. LoRFT instead provides manually verified observed/reference trajectory labels for the same vehicle within one fixed surveillance view, including distant portions that remain visually identifiable after automatic tracklets become fragmented or prematurely terminated.

\subsection{Vehicle Trajectory Forecasting}

Vehicle trajectory forecasting estimates future positions from observed motion histories. Early studies used kinematic models, maneuver recognition, probabilistic inference, hidden Markov models, or dynamic Bayesian networks to describe vehicle motion \cite{houenou2013motion,he2012hmm,schreier2016maneuver}. These methods are efficient and interpretable, but their simplified assumptions limit their ability to model nonlinear driving behavior and dense vehicle interactions. Recent surveys summarize this shift from model-driven prediction to learning-based trajectory forecasting \cite{bharilya2024survey}.

Deep models such as CS-LSTM, DeepTrack, GRIP++, and SCALE-Net learn temporal dynamics and vehicle interactions from trajectory histories~\cite{cs_lstm,deeptrack,grippp,scalenet}. Map-aware or physics-informed methods such as DenseTNT, LaneGCN, VectorNet, MixNet, and GNP further incorporate goal candidates, lane graphs, vectorized maps, or motion constraints \cite{densetnt,lanegcn,vectornet,mixnet,gnp2025}. ITPNet studies prediction with limited observations by recovering missing historical information before forecasting \cite{itpnet}.

These methods provide relevant trajectory modeling baselines, but their main task remains future prediction from usable observation histories. In LoRFT, the input is a reliable near-field tracklet from a fixed highway surveillance view, and the target is the missing distant segment of the same trajectory. The problem is therefore image-space long-range reconstruction of the same trajectory, rather than forecasting future motion from a complete and reliable observation history.

\subsection{Trajectory Reconstruction Under Incomplete Observations}

Trajectory imputation and reconstruction address incomplete observations more directly than standard forecasting. Xu et al. study trajectory imputation and prediction within a unified framework, showing that missing trajectory observations can be modeled as part of the learning problem~\cite{missing_pattern}. I-24 MOTION provides an instrumented freeway system for continuous vehicle trajectory collection~\cite{i24_motion}, and subsequent work on automatic trajectory reconstruction shows that converting imperfect visual detections into continuous trajectories is itself a substantive problem~\cite{wang2024reconstruction}. At a broader spatial scale, TrajRecovery recovers vehicle trajectories from urban traffic camera records by linking observations across a camera network~\cite{trajrecovery}. Recent roadside perception work further reconstructs full-sample trajectories from low-quality sensing data by combining data-driven representation learning with kinematic constraints~\cite{zeng2025roadside}.

These studies are closely related to our motivation, but they do not cover the single-camera near-to-far setting studied in LoRFT. Trajectory imputation typically recovers missing samples within an otherwise observed sequence, dense freeway systems rely on extensive camera coverage, city-scale recovery uses network-level traffic-camera evidence, and full-sample roadside reconstruction often refines low-quality detected trajectories. LoRFT instead considers long-range reconstruction within a single fixed highway surveillance view. The input is a reliable near-field tracklet, and the target is the distant continuation of the same vehicle after automatic tracking becomes unreliable. Map-RSTNet addresses this setting by using road geometry as the main structural prior for reconstruction under partial observability.

\begin{figure}[t]
\centering
\includegraphics[width=0.65\textwidth]{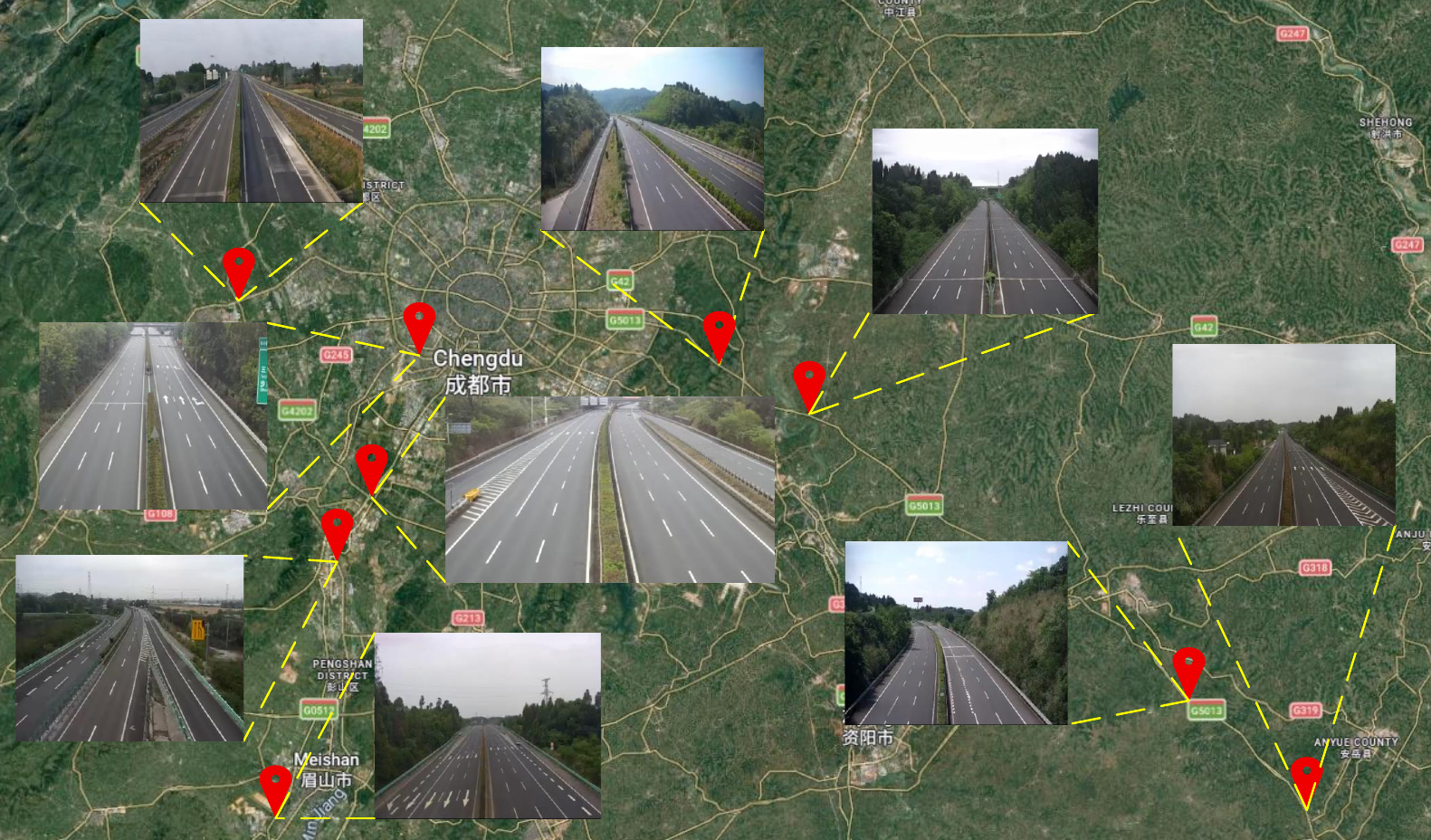}
\caption{Spatial distribution of selected LoRFT recording sites on expressways in Sichuan Province.}
\label{fig:dataset_sites}
\end{figure}

\section{LoRFT Dataset}
\label{sec:dataset}

\subsection{Data Collection and Scene Characteristics}

LoRFT is built for long-range vehicle trajectory reconstruction from fixed highway surveillance videos. The dataset was collected from 22 different sites on expressways in Sichuan Province, China, with videos recorded by roadside and gantry-mounted cameras. All videos are stored at \(352\times288\) pixels and 25 FPS.

As illustrated in Fig.~\ref{fig:dataset_sites}, the collected sites cover representative fixed-camera highway scenes, including straight mainline segments, curved segments, and merge or diverge areas. These scenes vary in road layout, traffic direction, camera viewpoint, and visible road extent. Because trajectories are represented in image coordinates, these scene-level differences affect the apparent displacement, scale variation, far-range visual clarity, and tracking continuity of near-to-far vehicle motion.

LoRFT focuses on the image-space degradation that occurs in fixed highway cameras. Vehicles near the camera usually have sufficient image scale for detection and tracking. As vehicles move into distant road regions, their apparent scale decreases, visual details become blurred, and identity association becomes less stable. Consequently, an automatically extracted tracklet may remain reliable near the camera but become fragmented or prematurely terminated while the vehicle is still visible in the same fixed-camera view. This observation motivates LoRFT's reconstruction protocol: the input is a reliable partial tracklet, and the target is the visually identifiable distant continuation of the same vehicle in the original image coordinate system.

\subsection{Trajectory and Road-Geometry Annotation}

The final LoRFT annotation set contains 366,109 video frames, 6,601 manually verified vehicle trajectories, and 2,694,889 vehicle bounding boxes. Eight trained annotators labeled and verified the collected videos, and multiple rounds of cross-checking were conducted to improve annotation consistency and reduce labeling errors. LoRFT adopts a semi-automatic annotation pipeline: candidate vehicle bounding boxes and tracklets are first obtained using a YOLOv11 detector and the ByteTrack tracker~\cite{bytetrack}. The detector-tracker output provides initial vehicle locations and identities, and reveals common fixed-camera detection and tracking failures, including missed detections, fragmented associations, identity switches, and early track termination in distant road regions. These automatically generated tracklets are used as annotation candidates rather than final labels.

The final vehicle trajectories are produced through frame-level manual verification. Annotators correct localization errors, association mistakes, false detections, and short occlusion-induced interruptions in the candidate tracklets. The observed segment denotes the portion where the detector-tracker output remains continuous and reliable after manual verification. In distant road regions, vehicles may be small or blurred in a single frame, but their identities can often be verified from their motion continuity across neighboring frames. If the same vehicle can still be confirmed after the automatic tracklet becomes unreliable or discontinuous, annotators extend or correct the long-range portion to form the image-space reconstruction reference. The split between the observed segment and the distant reference segment is therefore based on the reliability of the automatic tracklet, rather than on a fixed pixel threshold.

Each annotation row stores the frame index, vehicle identity, bounding box, and an observed/distant-segment label. Rows with \(\mathrm{label}=0\) denote the manually verified observed segment used as input in the reconstruction protocol, whereas rows with \(\mathrm{label}=1\) denote the manually verified distant reference segment used for image-space reconstruction evaluation. These labels specify the input and reference portions of the same fixed-camera trajectory; they do not necessarily indicate chronological order in the original video. Depending on traffic direction and camera placement, the distant reference segment may appear later or earlier in video time. LoRFT records trajectories in image coordinates because routine highway surveillance videos often lack reliable scene-specific camera calibration. The benchmark therefore evaluates reconstruction directly in the original image plane, without assuming complete metric trajectories or scene-specific camera calibration.

LoRFT also includes scene-level road-geometry annotations, referred to as scene maps in this paper. Each scene map contains road centerlines, left and right road boundaries, and zone-level traffic-direction metadata. The boundaries define the drivable region in the image plane, while the centerlines and zone metadata describe local road direction and traffic-flow structure. These annotations support road-aligned trajectory representation and geometry-aware reconstruction under fixed-camera surveillance views.

LoRFT further provides predefined scene-level train/validation/test splits, evaluation scripts, and model code. The scene-level split prevents trajectories from the same fixed surveillance view from appearing in multiple subsets, reducing the risk of overfitting to scene-specific road geometry, camera viewpoint, and background appearance. The released package includes vehicle-trajectory files, road-geometry annotations, split definitions, evaluation scripts, and model code.

\section{Method}
\label{sec:method}

\begin{figure}[t]
\centering
\includegraphics[width=\textwidth]{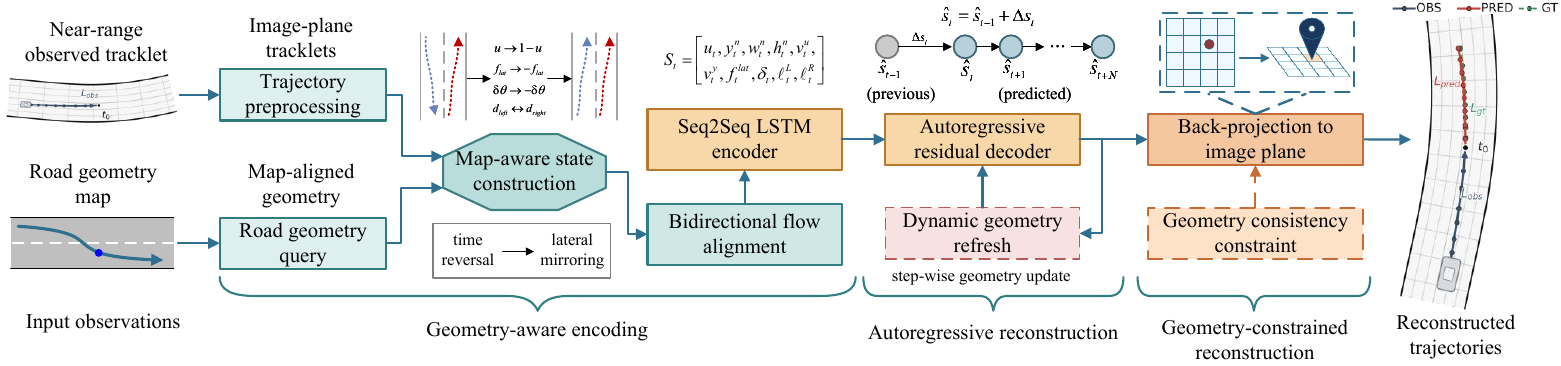}
\caption{Overall framework of Map-RSTNet. Near-field observed tracklets and scene-level road geometry are first encoded into a map-aware state representation. Bidirectional traffic flows are aligned into a unified motion space, and an autoregressive residual decoder reconstructs the missing segment while dynamically refreshing local road geometry. The reconstructed states are finally projected back to the image plane with geometry consistency constraints.}
\label{fig:framework}
\end{figure}

This section presents Map-RSTNet, a map-aware residual Seq2Seq LSTM framework for long-range trajectory reconstruction in fixed highway surveillance scenes. As shown in Fig.~\ref{fig:framework}, Map-RSTNet first converts image-space tracklets into a road-aligned state space using scene-level road geometry, then aligns bidirectional traffic into a shared motion space, and finally reconstructs the distant segment through autoregressive residual decoding with dynamic geometry refresh.

\subsection{Problem Formulation}

Given a fixed highway surveillance scene with a scene-level road prior \(\mathcal{M}\), Map-RSTNet reconstructs the distant segment of a vehicle trajectory from its reliable near-field observation. For frame \(t\), let the detected vehicle box be
\begin{equation}
    b_t=(x_t^b,y_t^b,w_t,h_t),
\end{equation}
where \((x_t^b,y_t^b)\) is the upper-left corner and \((w_t,h_t)\) is the box size. We use the bottom-center point as the image-space trajectory point:
\begin{equation}
    p_t=(x_t^{bc},y_t^{bc})
    =
    \left(x_t^b+\frac{w_t}{2},\;y_t^b+h_t\right),
\end{equation}
because it better approximates the vehicle-road contact location than the box center.

The observed input is a near-field box sequence \(\mathcal{O}_{1:T_o}\). Given \(\mathcal{O}_{1:T_o}\) and \(\mathcal{M}\), the model reconstructs \(T_p\) trajectory points of the same vehicle in image space. The reference segment is available during training and evaluation, but unavailable during inference. Since traffic direction and camera placement vary across scenes, Map-RSTNet normalizes each sample into an observed-to-distant reconstruction order and restores the reconstructed trajectory to the original image-space order for evaluation.

\subsection{Map-Aware State Encoding}

The road prior is represented as
\begin{equation}
    \mathcal{M}=\{\mathcal{B},\mathcal{C},\mathcal{Z}\},
\end{equation}
where \(\mathcal{B}\), \(\mathcal{C}\), and \(\mathcal{Z}\) denote annotated road boundaries, road centerlines, and zone-level traffic direction information, respectively. Before map matching, the bottom-center trajectory within each processed segment is smoothed with a constant-velocity Rauch--Tung--Striebel smoother to reduce frame-level localization jitter and obtain stable heading estimates. The preprocessing follows the reconstruction split: input states are constructed only from the observed segment, while the reference segment is processed separately and used only as supervision.

For each smoothed point \(p_t\), candidate centerlines are selected by spatial proximity and traffic direction. The final centerline is chosen by consistency between the vehicle heading \(\theta_t^{car}\) and the local road heading \(\theta_t^{road}\). The matched geometry provides the distances \(d_t^L\) and \(d_t^R\) from \(p_t\) to the left and right road boundaries, with local road width
\begin{equation}
    W_t=d_t^L+d_t^R .
\end{equation}

Each frame is encoded as a 10-dimensional road-aware state vector:
\begin{equation}
    s_t=
    [u_t,y_t^n,w_t^n,h_t^n,
    v_t^u,v_t^y,f_t^{lat},\delta_t,\ell_t^L,\ell_t^R].
\end{equation}
The position and scale terms are defined as
\begin{equation}
    u_t=\frac{d_t^L}{W_t},\quad
    y_t^n=\frac{y_t^{bc}}{H},\quad
    w_t^n=\frac{w_t}{W_t},\quad
    h_t^n=\frac{h_t}{W_t},
\end{equation}
where \(H\) is the image height. The velocity terms \(v_t^u\) and \(v_t^y\) are finite differences of \((u_t,y_t^n)\) with a fixed scaling factor. The geometry terms are
\begin{equation}
\begin{aligned}
    f_t^{lat} &= \frac{d_t^L-d_t^R}{W_t}, &
    \ell_t^L &= \frac{d_t^L}{W_t},\\
    \ell_t^R &= \frac{d_t^R}{W_t}, &
    \delta_t &= \operatorname{wrap}(\theta_t^{car}-\theta_t^{road}).
\end{aligned}
\end{equation}
This representation reduces perspective-related ambiguity by describing vehicle position, scale, motion, and road geometry in a shared normalized state space.

\subsection{Direction-Aligned Residual Decoder}

Fixed highway surveillance scenes may contain traffic moving in opposite directions. Map-RSTNet converts opposite-direction trajectories to a common reference orientation before sequence modeling. For a trajectory moving opposite to the reference direction, the state sequence is reversed in time, the normalized lateral coordinate is mirrored, the left and right boundary-distance terms are swapped, and the signs of \(f_t^{lat}\) and \(\delta_t\) are adjusted. Velocity terms are recomputed after this transformation, and the inverse transformation is applied before image-space evaluation.

After direction alignment, Map-RSTNet uses a relative position representation. Let \(\mathbf{P}_t=[u_t,y_t^n]^T\) denote the normalized position. The last observed position is used as the anchor:
\begin{equation}
    \tilde{\mathbf{P}}_t=\mathbf{P}_t-\mathbf{P}_{T_o}.
\end{equation}
This relative encoding reduces dependence on scene-specific absolute coordinates and makes the decoder focus on local motion increments.

The encoder summarizes the observed road-aligned state sequence into hidden and cell states \((h,c)\), and the decoder autoregressively predicts residual displacements in the normalized position space:
\begin{equation}
    \Delta\hat{\mathbf{P}}_t=\mathrm{Dec}(r_{t-1},h,c),
    \qquad
    \hat{\tilde{\mathbf{P}}}_t
    =
    \hat{\tilde{\mathbf{P}}}_{t-1}
    +
    \Delta\hat{\mathbf{P}}_t .
\end{equation}
Before accumulation, each coordinate of \(\Delta\hat{\mathbf{P}}_t\) is hard-clamped to \([-\delta_{\max},\delta_{\max}]\) in the normalized road-aligned state space. This per-coordinate step limit reduces unstable autoregressive jumps while preserving the residual decoding form. The decoder input \(r_{t-1}\) contains the reconstructed position, box-size terms from the last observation, velocity feedback from the residual, and the current geometry terms.

\subsection{Dynamic Geometry Refresh}

A single geometry vector from the last observed frame can become inaccurate as the reconstructed vehicle moves into distant road regions. Map-RSTNet therefore refreshes local geometry during decoding. At each decoding step, the predicted relative position is restored to the normalized road-aligned coordinate system,
\begin{equation}
    \hat{\mathbf{P}}_t
    =
    \hat{\tilde{\mathbf{P}}}_t+\mathbf{P}_{T_o},
\end{equation}
and then projected to an image-space query point using the latest valid boundary reference and road width. Let \(\bar{x}_{t-1}^{L}\) and \(\bar{W}_{t-1}\) denote the latest valid left-boundary reference and road width, and let \(\bar{u}_t\) be the lateral coordinate after inverse mirroring when direction alignment has been applied. The query point is computed as
\begin{equation}
    x_t^q=\bar{x}_{t-1}^{L}+\bar{u}_t\bar{W}_{t-1},
    \qquad
    y_t^q=\hat{y}_t^nH.
\end{equation}
The query point is matched to the scene road prior to update the local road width, boundary distances, lateral offset, and heading difference. If the map query is invalid, the previous valid geometry is retained.

This refresh mechanism keeps the autoregressive decoder conditioned on road geometry near the current reconstructed location rather than on outdated geometry from the observation endpoint. It is especially useful when road width, road heading, or perspective scale changes along the recovered trajectory.

\subsection{Objective and Coordinate Recovery}

Map-RSTNet is trained in the normalized road-aligned state space with a weighted objective:
\begin{equation}
    \mathcal{L}
    =
    \lambda_{pos}\mathcal{L}_{pos}
    +\lambda_{vel}\mathcal{L}_{vel}
    +\lambda_{end}\mathcal{L}_{end}
    +\lambda_{bdry}\mathcal{L}_{bdry}.
\end{equation}
The position term uses mean squared error in the normalized road-aligned position space:
\begin{equation}
    \mathcal{L}_{pos}
    =
    \frac{1}{T_p}
    \sum_{k=1}^{T_p}
    \left\|
    \hat{\mathbf{P}}_k-\mathbf{P}_k
    \right\|_2^2 .
\end{equation}
The velocity term uses a smooth-\(L_1\) penalty between residual-derived velocity and target velocity features, and the endpoint term penalizes the final-frame reconstruction error. The boundary term penalizes lateral predictions outside the normalized road interval:
\begin{equation}
    \mathcal{L}_{bdry}
    =
    \frac{1}{T_p}
    \sum_{k=1}^{T_p}
    \left[
    \max(0,-\hat{u}_k)^2
    +
    \max(0,\hat{u}_k-1)^2
    \right].
\end{equation}
The loss weights are selected on the validation set and kept fixed for all reported test experiments.

During inference, the anchor shift and direction-alignment transformation are inverted before coordinate recovery. Let \(x_k^L\) and \(W_k\) denote the left-boundary reference and refreshed road width used at step \(k\). The reconstructed bottom-center point is recovered as
\begin{equation}
    \hat{x}_k^{bc}=x_k^L+\hat{u}_k W_k,
    \qquad
    \hat{y}_k^{bc}=\hat{y}_k^n H.
\end{equation}
The recovered points are concatenated with the observed tracklet in the original temporal order to form the reconstructed image-space trajectory.

\section{Experiments}
\label{sec:experiments}

\subsection{Experimental Setup}

Map-RSTNet is evaluated on the proposed LoRFT benchmark. We use a scene-level split rather than a trajectory-level split. Specifically, the 22 fixed highway surveillance scenes are divided into 14 training scenes, 4 validation scenes, and 4 test scenes, approximately following a 60/20/20 ratio. This split is used because trajectories from the same surveillance view share similar road geometry, camera perspective, and background appearance. Separating scenes therefore provides a stricter test of generalization to unseen fixed-camera views.

For each annotated trajectory, we construct an input segment and a reference segment according to the LoRFT reconstruction protocol. The final setting uses 60 observed frames as input and a 125-frame distant reference segment as the reconstruction target. During training, sliding windows with a step of 5 frames are generated from eligible trajectories. During evaluation, only rows annotated with \(\mathrm{label}=0\) are provided as input, and the reconstructed segment is compared with rows annotated with \(\mathrm{label}=1\).

The validation set is used for model selection and hyperparameter tuning, and all reported results are computed on the held-out test scenes. Map-RSTNet is trained with Adam using a batch size of 128 and an initial learning rate of \(7\times10^{-5}\). The loss weights are set to \(\lambda_{pos}=1200\), \(\lambda_{vel}=10\), \(\lambda_{end}=150\), and \(\lambda_{bdry}=40\), with the velocity scaling factor \(\alpha_v=10\). Training is stopped according to validation performance. All experiments are conducted on an Ubuntu workstation with 64 GB RAM and two NVIDIA GeForce RTX 3090 GPUs.

\subsection{Evaluation Metrics}

\begin{table}[t]
\centering
\caption{Performance comparison on the LoRFT benchmark. Errors are measured in pixels; lower values are better.}
\label{tab:main_comparison}
\small
\begin{tabular}{lccccccc}
\hline
Model & ADE & FDE & RMSE@1s & RMSE@2s & RMSE@3s & RMSE@4s & RMSE@5s \\
\hline
CS-LSTM~\cite{cs_lstm}       & \underline{13.85} & 26.99 & \underline{8.48} & 16.42 & 25.12 & 35.72 & 48.87 \\
GRIP++~\cite{grippp}         & 17.50 & 33.80 & 8.77 & 16.14 & 23.25 & 30.65 & 38.57 \\
DeepTrack~\cite{deeptrack}   & 29.77 & 44.53 & 31.89 & 44.44 & 53.85 & 63.61 & 68.54 \\
MixNet~\cite{mixnet}         & 28.04 & 48.75 & 19.58 & 28.24 & 37.60 & 50.42 & 64.74 \\
GNP~\cite{gnp2025}           & 20.50 & 36.36 & 18.61 & 21.57 & 24.75 & 35.15 & 64.12 \\
PRF~\cite{zhou2026recover}   & 15.53 & \underline{25.67} & 12.20 & \underline{13.58} & \underline{18.15} & \underline{24.40} & \underline{30.70} \\
Map-RSTNet                   & \textbf{12.32} & \textbf{21.71} & \textbf{8.15} & \textbf{13.30} & \textbf{17.97} & \textbf{22.32} & \textbf{27.47} \\
\hline
\end{tabular}
\end{table}

Because LoRFT provides image-space vehicle trajectories from fixed surveillance videos without scene-level metric calibration, we use pixel-space errors as the primary evaluation metrics. After coordinate recovery, predictions are converted to the LoRFT image-coordinate format. Let \(\hat{\mathbf{a}}_{i,k}\) and \(\mathbf{a}_{i,k}\) denote the reconstructed and reference image-coordinate points of the \(i\)-th valid test trajectory at reconstruction step \(k\), respectively. We define the point-wise reconstruction error as
\begin{equation}
    e_{i,k}
    =
    \left\|
    \hat{\mathbf{a}}_{i,k}
    -
    \mathbf{a}_{i,k}
    \right\|_2 .
\end{equation}
Here, \(N\) is the number of valid test trajectories and \(T_p\) is the reconstruction horizon.

Average Displacement Error (ADE) measures the mean reconstruction error over the whole target segment:
\begin{equation}
    \mathrm{ADE}
    =
    \frac{1}{N T_p}
    \sum_{i=1}^{N}
    \sum_{k=1}^{T_p}
    e_{i,k}.
\end{equation}

Final Displacement Error (FDE) measures the endpoint error of the reconstructed segment:
\begin{equation}
    \mathrm{FDE}
    =
    \frac{1}{N}
    \sum_{i=1}^{N}
    e_{i,T_p}.
\end{equation}

We also report RMSE at fixed time horizons to evaluate error growth over the reconstruction period:
\begin{equation}
    \mathrm{RMSE}@\tau
    =
    \sqrt{
    \frac{1}{N_{\tau}}
    \sum_{i=1}^{N_{\tau}}
    e_{i,k_{\tau}}^2
    },
\end{equation}
where \(\tau\in\{1,2,3,4,5\}\) seconds, \(k_{\tau}=\tau f\), and \(f=25\) is the video frame rate. \(N_{\tau}\) denotes the number of valid trajectories that reach the corresponding horizon. Lower values indicate better reconstruction accuracy.

\subsection{Comparison with Representative Methods}

We compare Map-RSTNet with adapted trajectory forecasting and trajectory modeling baselines under the same LoRFT reconstruction protocol. The compared methods do not use lane ID as an input feature, and trajectories from both traffic directions are included in the evaluation. CS-LSTM~\cite{cs_lstm} is an encoder-decoder LSTM model with convolutional social pooling for interaction modeling. GRIP++~\cite{grippp} uses graph-based spatio-temporal interaction modeling. DeepTrack~\cite{deeptrack} is a lightweight highway trajectory prediction model designed for real-time inference. MixNet~\cite{mixnet} combines neural prediction with physics-based trajectory constraints. GNP~\cite{gnp2025} is a goal-based neural physics model that combines intention reasoning with neural motion generation. PRF~\cite{zhou2026recover} uses progressive retrospective learning to recover historical information for variable-length trajectory prediction.

As shown in Table~\ref{tab:main_comparison}, Map-RSTNet achieves the lowest error on all metrics. Compared with the best baseline for each metric, Map-RSTNet reduces ADE from 13.85 to 12.32, FDE from 25.67 to 21.71, and RMSE@5s from 30.70 to 27.47. These correspond to relative reductions of 11.0\%, 15.4\%, and 10.5\%, respectively. PRF narrows the gap at intermediate horizons, especially at 2s and 3s, but Map-RSTNet remains consistently better across the full reconstruction horizon. The larger gains at 4s and 5s suggest that road-aligned representation and dynamic geometry refresh help limit error accumulation during long-range reconstruction in fixed highway surveillance scenes.

\begin{figure}[t]
\centering
\includegraphics[width=0.95\textwidth]{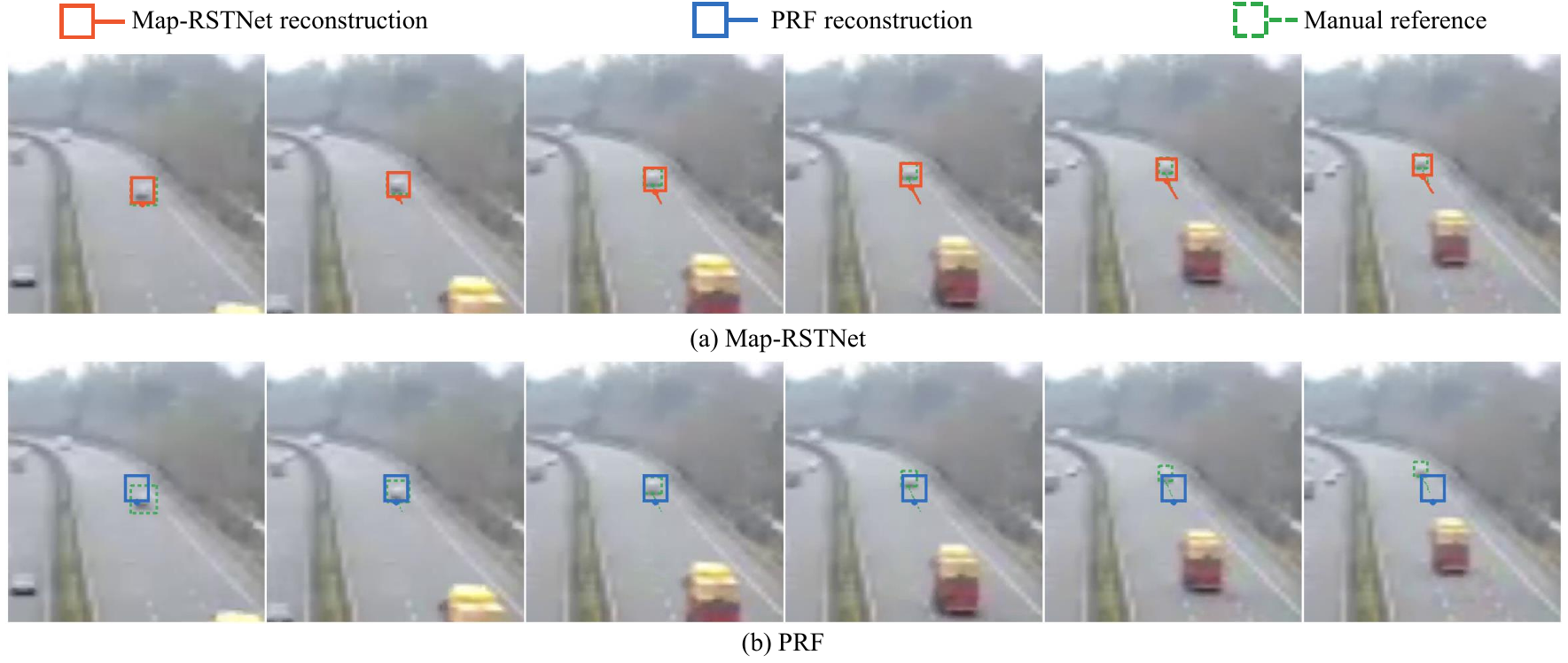}
\caption{Qualitative comparison between Map-RSTNet and PRF on a LoRFT test scene. 
The sampled frames show the same far-range vehicle during reconstruction. 
Orange and blue boxes with short traces denote the reconstructed results of Map-RSTNet and PRF, respectively, while green boxes denote the manually verified reference. 
Map-RSTNet remains closer to the reference as the vehicle moves through the degraded far-range region.}
\label{fig:qualitative_comparison}
\end{figure}

Fig.~\ref{fig:qualitative_comparison} provides a qualitative comparison with PRF, the strongest baseline in the long-horizon metrics. In this far-range example, the vehicle is small and blurred in individual frames, but the manually verified reference remains traceable from temporal context. Map-RSTNet stays closer to the reference across the sampled frames, whereas PRF shows larger drift as the vehicle moves through the distant road region.

\subsection{Ablation Study and Analysis}

\begin{table}[t]
\centering
\caption{Ablation study on the LoRFT benchmark. Errors are measured in pixels; lower values are better.}
\label{tab:ablation}
\renewcommand{\arraystretch}{1.12}
\begin{tabular*}{\columnwidth}{@{\extracolsep{\fill}}cccccc}
\hline
Mapping & Offset & Refresh & ADE & FDE & RMSE@5s \\
\hline
        &        &        & 15.46 & 28.04 & 32.92 \\
\(\checkmark\) &        &        & 14.20 & 26.59 & 31.78 \\
\(\checkmark\) & \(\checkmark\) &        & 13.97 & 25.45 & 30.26 \\
\(\checkmark\) & \(\checkmark\) & \(\checkmark\) & \textbf{12.32} & \textbf{21.71} & \textbf{27.47} \\
\hline
\end{tabular*}
\end{table}

We conduct a cumulative ablation study to examine three components of Map-RSTNet: map-aware state encoding, local offset encoding, and dynamic geometry refresh. The first row keeps the same residual Seq2Seq backbone without these three components. The following rows progressively add the proposed designs under the same LoRFT reconstruction protocol. Table~\ref{tab:ablation} reports the results.

In Table~\ref{tab:ablation}, Map, Offset, and Refresh refer to map-aware state encoding, local offset encoding, and dynamic geometry refresh, respectively. The ablation is cumulative: starting from the residual Seq2Seq backbone, the three components are added step by step. All errors are measured in pixels, and lower values indicate better reconstruction accuracy.

The first two rows evaluate the effect of introducing road geometry into the trajectory representation. After map-aware state encoding is added, ADE decreases from 15.46 to 14.20, FDE from 28.04 to 26.59, and RMSE@5s from 32.92 to 31.78. This improvement suggests that raw image coordinates alone are insufficient for fixed highway surveillance videos, where the same vehicle motion can correspond to different pixel displacements at different depths. Encoding trajectories with respect to local road geometry gives the model a more stable reconstruction space.

The next comparison examines local offset encoding. Its effect on ADE is modest, but the endpoint and long-horizon errors are reduced more clearly, with FDE decreasing from 26.59 to 25.45 and RMSE@5s from 31.78 to 30.26. This trend matches the role of the relative position design: anchoring the sequence at the last observed state helps the decoder focus on local residual motion instead of fitting scene-specific absolute positions.

The final row shows the contribution of dynamic geometry refresh. Refreshing the road-geometry features during decoding reduces ADE from 13.97 to 12.32, FDE from 25.45 to 21.71, and RMSE@5s from 30.26 to 27.47. The larger reductions in FDE and RMSE@5s indicate that this component mainly improves long-range stability. This is consistent with the reconstruction setting of LoRFT, where the predicted vehicle moves into distant road regions and a fixed geometry vector from the observation endpoint can become inaccurate.

\begin{figure}[t]
\centering
\includegraphics[width=0.92\textwidth]{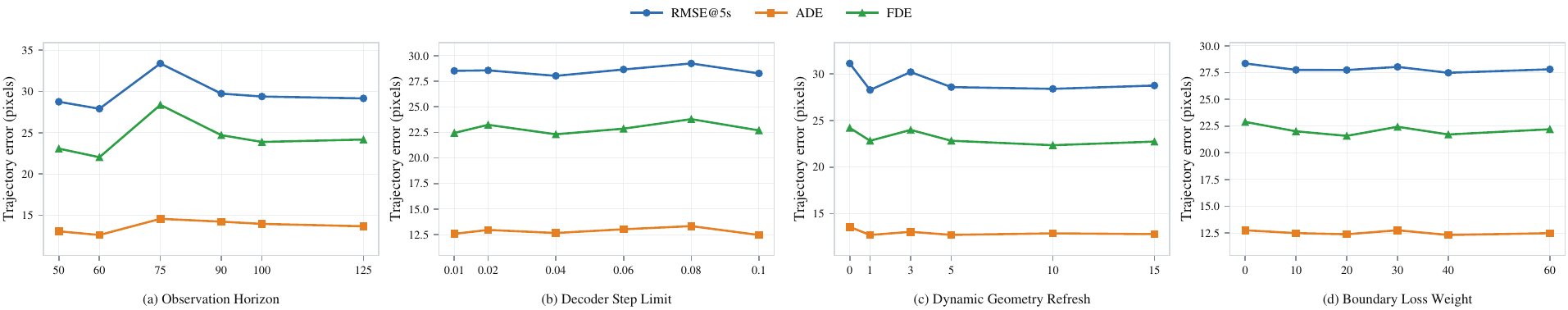}
\caption{Impact of key hyperparameters on the LoRFT benchmark. (a) Observation horizon. (b) Decoder step limit. (c) Dynamic geometry refresh interval. (d) Boundary loss weight. Errors are measured in pixels; lower values are better.}
\label{fig:hyperparameters}
\end{figure}

\subsection{Impact of Hyperparameters}

We further analyze the sensitivity of Map-RSTNet to four hyperparameters: the observation horizon, the decoder step limit, the dynamic geometry refresh interval, and the boundary loss weight. The final configuration is selected according to validation performance and then kept fixed for the test evaluation.

Fig.~\ref{fig:hyperparameters}(a) shows the effect of the observation horizon. A short observation window provides limited motion context, making it difficult to infer the long-range continuation of a partially observed trajectory. Increasing the observation length initially improves reconstruction, but overly long histories do not bring further gains. In fixed highway surveillance videos, longer observations may include redundant temporal information and perspective-induced noise from the near-field tracklet. The best performance is obtained with a 60-frame observation window, which is used in the final model.

Fig.~\ref{fig:hyperparameters}(b) evaluates the decoder step limit. The decoder step limit denotes \(\delta_{\max}\), the per-coordinate hard-clamp threshold applied to each predicted residual in the normalized road-aligned state space. A small limit restricts the decoder and may prevent it from following valid motion changes, while a large limit weakens the regularization effect and can increase drift over long horizons. The model remains relatively stable over a moderate range of values, and the best validation performance is achieved when the step limit is set to 0.04.

Fig.~\ref{fig:hyperparameters}(c) examines the refresh interval used by dynamic geometry refresh. Disabling refresh gives the weakest performance, which indicates that the geometry at the last observed frame is insufficient for long-range reconstruction. Refreshing geometry more frequently allows the decoder to use road width, boundary distance, and heading information that better matches the current reconstructed location. The best result is obtained when geometry is refreshed at every decoding step, so the final model uses a refresh interval of one frame.

Fig.~\ref{fig:hyperparameters}(d) reports the effect of the boundary loss weight. The model is less sensitive to this parameter than to the observation horizon or geometry refresh interval, but removing or underweighting the boundary term weakens the road feasibility constraint. A very large weight can overemphasize boundary adherence and slightly interfere with motion reconstruction. The best overall validation performance is obtained with a boundary loss weight of 40, which is used in all reported experiments.

\section{Application Scenarios}
\label{sec:application}

LoRFT is intended for fixed-camera highway settings where perception quality changes substantially with distance. Beyond the reconstruction protocol studied in this paper, LoRFT can support evaluation and analysis of far-range vehicle detection and tracking in fixed highway views, because it provides raw surveillance videos, frame-level boxes with identities, manually verified trajectories, and scene-level road geometry. These annotations make it possible to examine detection misses, identity switches, track fragmentation, and premature termination under the same surveillance views used for reconstruction. The road-boundary and centerline annotations further support perspective-aware trajectory representations, where local road geometry helps reduce ambiguity from scale decay and perspective distortion.

Reconstructed trajectories can also serve as longer motion records for downstream traffic mining. In fixed highway surveillance, fragmented tracklets restrict analyses that depend on longer motion histories, such as trajectory prediction, lane-change and conflict analysis, active risk assessment, and data-driven traffic management. LoRFT provides a controlled setting for measuring how much trajectory continuity can be recovered before such downstream analyses are applied. These uses should be interpreted within the coordinate system of the dataset. LoRFT provides image-space trajectories and road-geometry annotations, but not scene-specific camera calibration parameters. Applications requiring metric speed, acceleration, or physical vehicle dynamics therefore need additional calibration or image-to-road mapping.

\section{Conclusion}
\label{sec:conclusion}

This paper studied long-range vehicle trajectory reconstruction from fixed highway surveillance videos, where reliable near-field tracklets often lose continuity in distant road regions. We introduced LoRFT, an open benchmark and evaluation protocol that provides manually verified observed/reference trajectory pairs, 6,601 vehicle trajectories from 366,109 video frames, 2,694,889 vehicle bounding boxes, and scene-level road-geometry annotations. We also proposed Map-RSTNet, a map-aware residual Seq2Seq model for road-aligned trajectory reconstruction.

Experiments on the scene-level LoRFT protocol show that Map-RSTNet reduces long-horizon reconstruction errors compared with adapted forecasting and trajectory modeling baselines. The ablation results further indicate that road-aligned state encoding, local offset modeling, and dynamic geometry refresh help control error accumulation over longer reconstruction horizons. These results suggest that fixed highway cameras can provide longer usable trajectory records when far-range track fragmentation is treated as a reconstruction problem.

\section*{Acknowledgment}
This work was supported by the National Natural Science Foundation of China under Grant No. 52411540233 and the Graduate Research and Innovation Project of Changsha University of Science and Technology under Grant No. CLKYCX25007.

\section*{Data and Code Availability}
Processed trajectory annotations, road-geometry files, evaluation scripts, and source code are publicly available at \url{https://github.com/YvfanZhu/LoRFT}. The original surveillance videos are distributed separately under controlled access; access instructions and the data-use agreement are provided in the repository.

\bibliographystyle{IEEEtran}
\bibliography{references}

\end{document}